\documentclass{svproc}
\usepackage{cite}
\usepackage{amsmath,amssymb,amsfonts}
\usepackage{algorithmic}
\usepackage{graphicx}
\usepackage{textcomp}
\usepackage{xcolor}
\usepackage[utf8]{inputenc}
\usepackage[english]{babel}

\usepackage{multirow}
\usepackage[misc]{ifsym}
\usepackage{cleveref}
\usepackage{url}

\begin{document}
\mainmatter              % start of a contribution
\title{Domain Generalization using Ensemble Learning}
%
%\titlerunning{Hamiltonian Mechanics}  % abbreviated title (for running head)
% Anonymous Submission}
\author{Yusuf Mesbah(\Letter),  Youssef Youssry Ibrahim, \and
Adil Mehood Khan}
\authorrunning{Yusuf Mesbah et al.} % abbreviated author list (for running head)

%%% list of authors for the TOC (use if author list has to be modified)
\tocauthor{Ivar Ekeland, Roger Temam, Jeffrey Dean, David Grove,
Craig Chambers, Kim B. Bruce, and Elisa Bertino}

\institute{Machine Learning and Knowledge Representation Lab, Innopolis University, Republic of Tatarstan, Russian Federation,\\
\email{y.mesbah@innopolis.university},
% \\ WWW home page:
% \texttt{http://users/\homedir iekeland/web/welcome.html}
% \and
% Universit\'{e} de Paris-Sud,
% Laboratoire d'Analyse Num\'{e}rique, B\^{a}timent 425,\\
% F-91405 Orsay Cedex, France
}

\maketitle              % typeset the title of the contribution

\begin{abstract}
Domain generalization is a sub-field of transfer learning that aims at bridging the gap between two different domains in the absence of any knowledge about the target domain. Our approach tackles the problem of a model's weak generalization when it is trained on a single source domain. From this perspective, we build an ensemble model on top of base deep learning models trained on a single source to enhance the generalization of their collective prediction. The results achieved thus far have demonstrated promising improvements of the ensemble over any of its base learners.
% We would like to encourage you to list your keywords within
% the abstract section using the \keywords{...} command.
\end{abstract}
\section{Introduction}
% ---------

Ensemble learning is a method in supervised learning that combines multiple predictive models to get better and more robust predictions, which makes ensemble learning methods the best choice when the performance is of high importance. When it comes to the number of classifiers in the ensemble, the work done by R. Bonab, Hamed; Can, Fazli (2016) demonstrating \textit{the law of diminishing returns in ensemble construction} can be referred to. Their theoretical framework shows that the highest accuracy is achieved by using the same number of independent component classifiers as class labels\cite{inproceedings}.
% ---------

% P2: Neural networks (deep learning) and their success in solving various challenging problems
The theoretical base of neural networks was proposed by Alexander Bain (1873) and William James (1890) independently. Later on, McCulloch and Pitts (1943) made a mathematical model based on neural networks and called it "threshold logic". After that, back-propagation was introduced by Rumelhart, Hinton, and Williams (1986). Over the following years, with the scientific and technological advancements, neural networks algorithms became more sophisticated and able to solve bigger and more challenging problems, including object recognition\cite{khan2020post}, anomaly detection\cite{rivera2020anomaly, yakovlev2020abstraction}, accident detection\cite{batanina2019domain, bortnikov2019accident}, action recognition\cite{gavrilin2019across, sozykin2018multi, khan2010accelerometer}, scene classification\cite{protasov2018using}, hyperspectral image classification\cite{ahmad2020fast, ahmad2019multi}, machine translation\cite{khusainova2019sart, valeev2019application}, medical image analysis\cite{gusarev2017deep, dobrenkii2017large}, etc. 
%neural networks algorithms became more sophisticated and able to solve bigger and more challenging problems.
%Nowadays, artificial neural networks are widely used in many industries for their efficiency and robustness, and in some cases even outperform humans, such as in image recognition, video games, and predictions \cite{sabour2017dynamic, mnih2015humanlevel, gebru2017using}.

\begin{figure}[t!]
    \centering
    \includegraphics[width=0.49\textwidth]{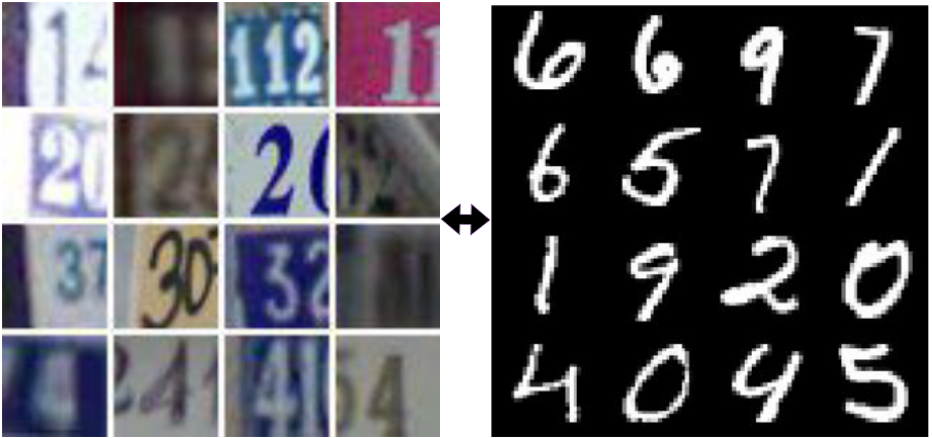}
    \caption{Domain generalization is the problem of transferring the knowledge from a source domain (such as SVHN\_cropped on the left) to a different target domain (such as MNIST on the right) to solve the same task, with the absence of any knowledge regarding the distributional shift in the feature space of the inputs.}
  \label{fig:SVHN and MNIST}
  
\end{figure}

Nowadays, deep learning (DL) and convolutional neural networks (CNN) are widely used in our everyday life. For example, modern smartphones have an option of authenticating using facial recognition, and all new self-driven cars are based mainly on a combination of many Deep CNNs to process road images. This increase in use raises the bar for computer vision systems to be more robust and stable. As useful as DL techniques are, some problems are faced when deploying them in the real world that we do not commonly encounter while working on toy datasets or training data in general. As powerful as deep CNNs are, they have a considerable shortcoming in that they are heavily dependant on the dataset used for training; this problem is also known as over-fitting.
The problem at hand (called domain-shift) is mainly due to the fact that the training data set (source domain) comes from a different distribution than the deployment data (target dataset), resulting in a decrease in the model's performance. Such discrepancy can occur in real life from slight changes in variables such as image resolution or picture brightness.
% ---------

% P3: Domain Generalization (DG), its importance and challenges
Domain Generalization (DG) is a sub-field of Transfer Learning (TL) that aims to solve this problem by combining multiple data sources to train a more resilient model in hopes of generalizing to unseen domains. DG assumes the existence of multiple sources of data that are used for the same task, and a target domain dataset that is harder to work with (i.e.: harder to label and/or to collect). All domains share the same task but have a different marginal distribution. DG is very closely related to Domain Adaptation (DA) which also aims at solving the domain shift problem using one source domain and one target domain. DA can be approached in different ways regarding the existence of labels in the target domain: Supervised, Unsupervised, or Semi-Supervised. DG differs from DA in the fact that we do not have access to the target data nor to its labels. So, Domain Generalization aims at building a model that can generalize well to unseen domains rather than generalizing to a single known domain. Researchers have approached this problem in many ways. One traditional - yet very commonly used - technique is to treat this problem as an over-fitting problem and use regularisation techniques to help the (parametric) model generalize well\cite{zhang2017understanding}. Many techniques have proven useful in the case of deep neural networks such as weight decay, dropout, batch normalization, and $L_1$  and $L_2$ regularization. Although these techniques were proven effective to help the model generalize well within the same data set and achieve higher test accuracy, they are not the most effective methods for Domain Generalization.
% ---------

% P4: Aim of this paper: (In general) performance of ensemble models for DG, (in particular) performance of ensemble of neural networks (ANN-En) for DG and its comparison (accuracy and training time) to a single deep network and ensemble of traditional machine learning methods like SVM, RF, LR, etc.
In this paper, we deal with the case of Domain Generalization in its largest definition, where we handle the case of generalizing from a source domain to an unknown target domain. More specifically, we compare the performance of ensemble models with an individual Deep Neural Network on a single source domain generalization. Since ensemble models have shown an increase in accuracy in difficult learning scenarios, we will be investigating how much benefit can ensemble models give us when dealing with Domain Generalization problems. Accordingly, in this paper, we have implemented various ensemble models that consist of CNNs and different traditional machine learning models. We have tested them on five different datasets, and are reporting very interesting findings.
% ---------

\section{Related Work}
% ---------

% subsection: Works that have used ensemble of ANNs
\subsection{Ensemble Learning}
Ensemble methods have been extensively researched. The main idea is to train multiple predictors for the same problem and merge their output to get better results. Ensemble methods have been commonly used in competitive machine learning competitions such as ILSVRC, where many CNNs are trained and merged to improve performance \cite{krizhevsky2012imagenet, mancini2018best, zhou2012ensemble}. One main difference between the traditional model ensembling and our approach (when it comes to hyperparameter tuning) is the size of the models, in that even though we can use bigger models that will have a better performance on the training data set (and subsequently an ensemble of them), we preferred weaker models that still perform well on the training dataset (0.9+ accuracy) while providing much better generalization (more on that in \Cref{experiments}).
% ---------

% subsection: Works on Transfer Learning
\subsection{Transfer Learning}
Transfer learning (TL) in machine learning is the topic that explores how to store and apply the knowledge gained while solving one problem in a different but related problem. For example, the knowledge gained about recognizing cars could apply when trying to recognize trucks\cite{west2007spring}. This is useful to decrease the training time of the models and helps if the target dataset is small. Similarly, Semi-Supervised classification\cite{yin2006efficient,kuncheva2007classifier, baralis2007lazy} tackles the problem of the labelled data not being large enough to build a strong classifier, utilizing the large amount of data and the small number of labels. For example, Zhu and Wu \cite{zhu2006class} discussed how to deal with noisy labels, and Yang et al. considered cost-sensitive learning \cite{yang2006test}. Semi-Supervised classification assumes that the distributions of the labelled and unlabelled data are from the same domain, while Transfer Learning allows the domains and tasks used in training and testing to be different\cite{5288526}.

% subsection: Works on DG
\subsection{Domain Generalization}

Domain Generalization is less explored as a topic than Domain Adaptation \cite{tripnet}, even though the ability to access multiple source domains allowed for more innovative and creative techniques. These techniques mainly fall into two streams:
\renewcommand{\labelenumi}{\roman{enumi}}
\begin{enumerate}
\item Combining the source domains in a way that helps the model learn domain invariant features that can generalize well to unseen domains. For example, one state of the art method tries to learn domain-agnostic representations by re-arranging the input images and asking the network to solve it as a jigsaw puzzle. Although it has proven very effective, it faces a risk when different classes can share the same sub-components but are linked together differently.
\item Measuring the similarity between each target image and potential source domains and then using this information, later on, to either combine or choose a certain classifier to use for this sample as in BSF \cite{mancini2018best}.
\end{enumerate}
% ---------
\section{Methods}
% ---------
\begin{figure*}
  \includegraphics[width=\textwidth]{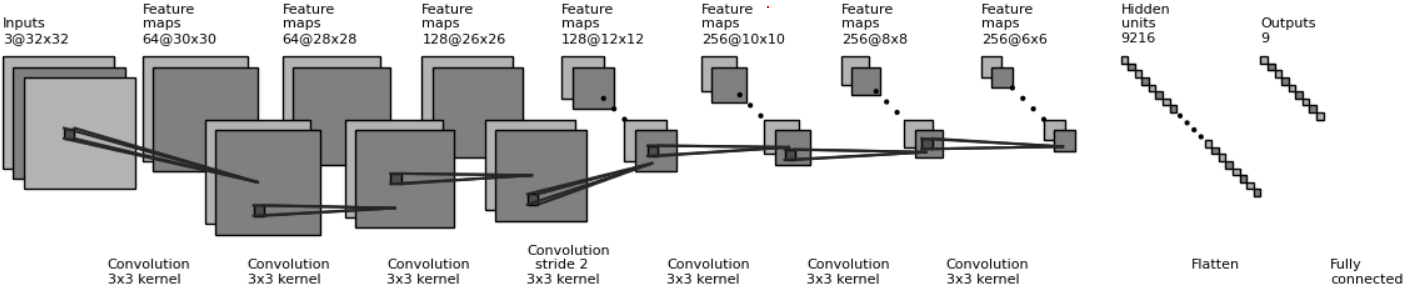}
  \caption{Base CNN used to learn CIFAR10 for the ensembles}
  \label{fig:CNN}
\end{figure*}
% Details of three variants of the ANN-En with their diagrams (use tikz for figures)

We will be comparing three different ensembles with a single Neural Network to evaluate which one performs better on different various generalization problems.

In supervised machine learning, there is some dataset $D$ that consists of input data points, where every data point denoted by $x$ has a class label $y$, with the assumption that there exists a function $f$ that maps from the data point to the class label as $y = f(x)$. The purpose of learners is to search through a space of possible functions, called hypotheses, to find the function $h$ which is the best approximation to $f$ used to assign the label $y$ to $x$. Such a function is called a classifier.

Learners that use a single hypothesis approximation for predictions could suffer from three main problems \cite{dietterich2002ensemble}:

\renewcommand{\labelenumi}{\roman{enumi}}
\begin{enumerate}
\item The \textbf{statistical} problem is when the learner is searching in a space of hypotheses that is too big for how much training data is available. In this case, there might be two or more hypotheses that get the same accuracy on the training data but perform differently while predicting future data. An ensemble can reduce the risk of this problem by taking the vote of different learners with different hypotheses, as it reduces the overall variance. In \cite{58871}, the authors illustrated the variance reduction property of an ensemble system.

\item The \textbf{computational} problem is when the learner is not guaranteed to find the best hypothesis and can get stuck in a local minimum as is the case with neural networks and decision tree algorithms. However, as with the statistical problem, an ensemble can help mitigate the computational problem because the weighted combination of several different local minima can help avoid choosing the wrong local minimum.

\item The \textbf{representational} problem is when the hypothesis space does not contain a good approximation of the true function $f$. An ensemble can help in some cases, as a weighted vote of the hypotheses can expand the hypothesis space and result in a better approximation of $f$ .
\end{enumerate}

The aforementioned problems can become even more severe when there is a domain gap between the training (source) data and the test (target) data. Usually, this problem is alleviated by training a model on multiple, different source domains. However, if there is a single domain to learn from, generalization could become extremely difficult. Therefore, it is interesting to see whether an ensemble model that uses a single source domain but benefits from having different base learners could help in improving the generalization performance. If so, what kind of ensemble model would perform better? Accordingly, our experiments are tailored to figure out answers to these questions.

For every experiment conducted in our paper, we will have a single source dataset $D^s$ and a single target dataset $D^t$ that has a different domain, then we will have $N$ CNN models (similar to Figure \ref{fig:CNN}) $(m_1, m_2, ..., m_N)$ that will be trained independently on $D^s$. Then, they will be tested on the target domain $D^t$, give us their predictions $(y_1', y_2', ..., y_N')$, then by getting the average of their output $\bar{y}'$, we get our first ensemble (average ensemble, denoted by EnA).

\begin{figure}[ht]
    \centering
    \includegraphics[width=0.4\textwidth]{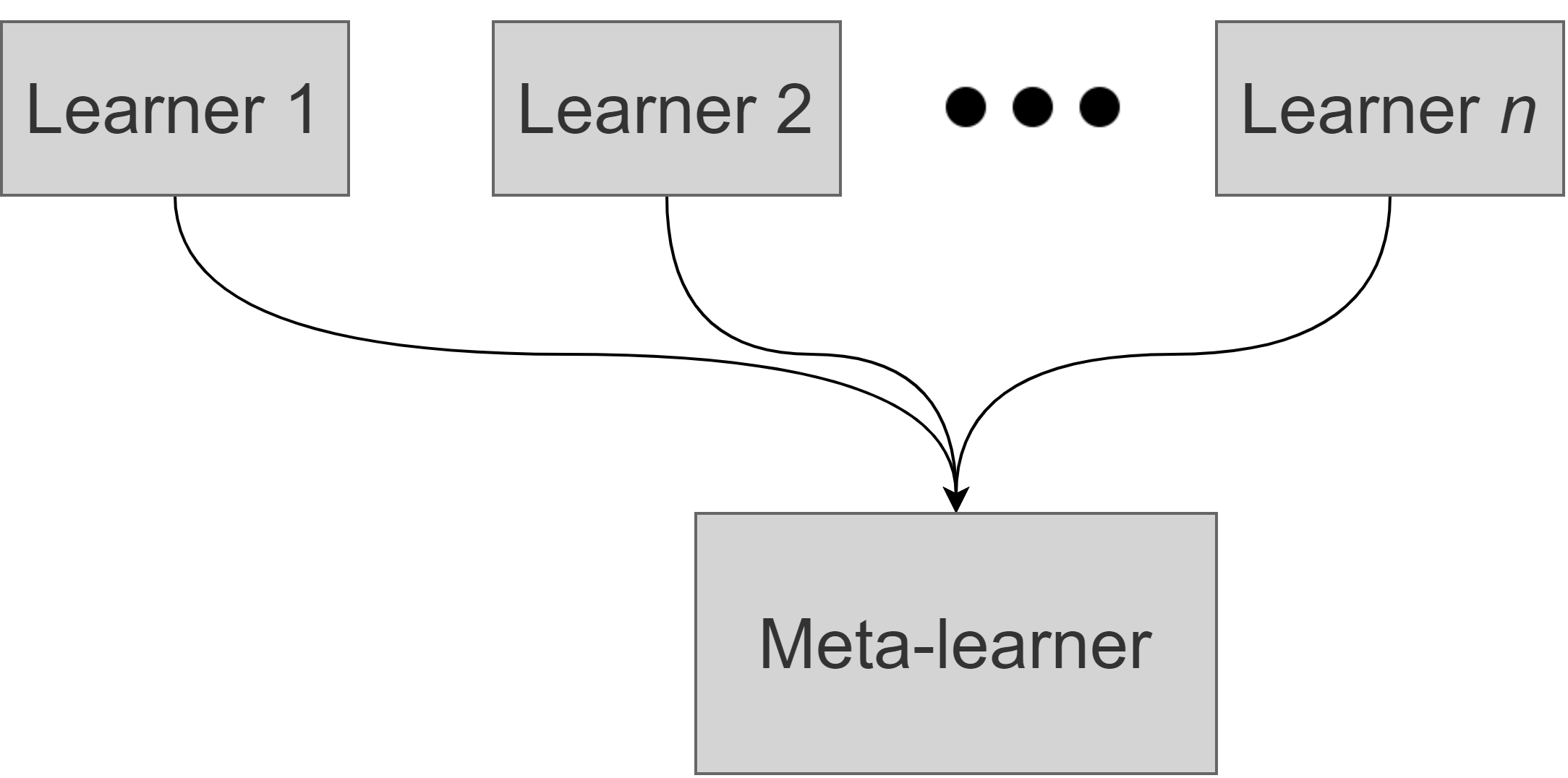}
    \caption{General ensemble model}
  \label{fig:general ensemble}
\end{figure}

For the second ensemble with the meta learner (EnM), we will take the models' outputs and train a layer of perceptrons as a meta learner to give us a weighted average of the models' outputs. See Figure \ref{fig:general ensemble}.

For the third ensemble, which is with meta learner v2 (EnM2), it will be similar to the previous ensemble with the only difference being that it has a multi layer perceptron meta learner. 

For the last ensemble, we compose different traditional ML algorithms (Random Forest (RF), Support Vector Machines (SVM), and Logistic Regression (LR)) into an average ensemble (EnT), see Figure \ref{fig:general ensemble}.

Lastly, we will be adding to the comparison a single huge CNN (HCNN) that has as much trainable parameters as the sum of all the CNNs in the ensemble to see how the different use of trainable parameters might affect the results.
% ---------

% ---------

% subsection: Data Preparation (Problem 1 and Problem 2)
\subsection{Data Preparation}
There will be two datasets from different domains; one of them will be used for training and hyperparameter-tuning, and the other will be for testing to see how the ensemble will perform on a different domain.

As for data preparation, for every neural network in the ensemble, a different data augmentation technique will be applied to the training dataset to increase the variance in the training data for every network.

% ---------

% subsection: Experiments
\subsection{Experiments}
\label{experiments}
For the first experiment, we will be using three digits datasets: MNIST\cite{lecun-mnisthandwrittendigit-2010}, USPS\cite{uspsdataset} and SVHN\_cropped \cite{netzer2011reading} (henceforth referred to as SVHN). MNIST and USPS are composed of white handwritten digits on a black background, but USPS is small and zoomed to fill the frame, while MNIST is large and padded. On the other hand, SVHN is composed of colored images on a colored background (see Figure \ref{fig:SVHN and MNIST}). Moreover, the digits in SVHN are not perfectly isolated; there can be more than one digit in the one image, and the label for this image would be the middle digit in the image. We will train on one dataset and test on another (for every possible pairing of the 3 datasets).

For the second experiment, we will use natural objects datasets CIFAR10\cite{cifar10} and STL10\cite{coates2011analysis}. CIFAR10 is a colored dataset that consists of 10 natural objects: 5 animals, and 5 vehicles. Similarly, STL10 has the same setup except that CIFAR10 has images of frogs and STL10 does not. On the other hand, STL10 has images of monkeys while CIFAR10 does not, so we removed the uncommon labels, leaving us with 9 labels in common between the 2 datasets.
In experiments involving USPS, the other datasets were resized to $16\times16$ to match USPS. In all other experiments, all the datasets were re-scaled to be $32\times32$ pixels. SVHN was converted to gray-scale to match MNIST \cite{dietterich2002ensemble}.
% ---------
% subsection: Hyperparameters and their tuning

\subsection{Hyperparameter tuning}
For every experiment that was done there were two datasets: source $S$ and target $T$ datasets. The source dataset is further divided into two parts: \textit{train} and \textit{validation}, so we will call them $S_{train}$ and $S_{val}$, respectively.

\subsubsection{CNNs and Ensemble Meta Classifier}
To train each CNN (Figure \ref{fig:CNN}), we used $S_{train}$ for training and $S_{val}$ for validation. To achieve independence between the base models, we have a set of different types of augmentations $A = \{a_1, a_2, ..., a_n\}$, and every model $i$ in the ensemble is trained using a unique subset of augmentations $A_i \subset A$.
On the other hand, the ensemble meta classifier and the single CNN that will have the same number of parameters as the ensemble were trained using the full set of augmentations $A$.

\subsubsection{Traditional ML Algorithms}
Similarly, we used $S_{train}$ for training and $S_{val}$ to tune some parameters such as the number of trees in a random forest.
\begin{table}[h!]
\centering
\begin{tabular}{|c|c|c|c|c|c|c|}
\hline
Model         & \multicolumn{3}{c|}{CIFAR10 to STL10}          & \multicolumn{3}{c|}{STL10 to CIFAR10}          \\ \hline
              & $S_{train}$  & $S_{val}$      & $T$            & $S_{train}$  & $S_{val}$      & $T$            \\ \hline
model 1       & 0.987        & 0.886          & 0.706          & 0.721        & 0.597          & 0.460          \\ \hline
model 2       & 0.978        & 0.879          & 0.675          & 0.944        & 0.664          & 0.557          \\ \hline
model 3       & 0.978        & 0.877          & 0.686          & 0.903        & 0.636          & 0.504          \\ \hline
model 4       & 0.976        & 0.868          & 0.684          & 0.984        & 0.641          & 0.509          \\ \hline
model 5       & 0.969        & 0.888          & 0.696          & 0.818        & 0.633          & 0.515          \\ \hline
\textbf{EnA}  & 0.99         & 0.903          & 0.724          & 0.964        & 0.681          & 0.558          \\ \hline
\textbf{EnM}  & 0.99         & \textbf{0.904} & \textbf{0.727} & 0.964        & \textbf{0.684} & 0.559          \\ \hline
\textbf{EnM2} & 0.99         & 0.903          & 0.725          & 0.973        & 0.68           & \textbf{0.563} \\ \hline
\textbf{HCNN} & 0.971        & 0.878          & 0.683          & 0.466        & 0.423          & 0.358          \\ \hline
\textbf{EnT}  & 0.958        & 0.487          & 0.366          & 0.709        & 0.371          & 0.285          \\ \hline
RF            & \textbf{1.0} & 0.498          & 0.373          & \textbf{1.0} & 0.459          & 0.305          \\ \hline
SVM           & 0.081        & 0.077          & 0.091          & 0.193        & 0.184          & 0.177          \\ \hline
LR            & 0.464        & 0.429          & 0.305          & 0.654        & 0.359          & 0.281          \\ \hline
\end{tabular}
\caption{Results for object recognition experiments: (1) from CIFAR10 as the source domain to STL10 as the target domain, (2) from STL10 as the source domain to CIFAR10 as the target domain.}
\label{table1}
\end{table}

\begin{table}[h!]
\centering
\begin{tabular}{|c|c|c|c|c|c|c|}
\hline
Model         & \multicolumn{3}{c|}{MNIST to SVHN}             & \multicolumn{3}{c|}{SVHN to MNIST}             \\ \hline
              & $S_{train}$  & $S_{val}$      & $T$            & $S_{train}$  & $S_{val}$      & $T$            \\ \hline
model 1       & 0.971        & 0.973          & 0.069          & 0.934        & \textbf{0.936} & 0.647          \\ \hline
model 2       & 0.971        & 0.973          & 0.069          & 0.933        & 0.935          & 0.648          \\ \hline
model 3       & 0.978        & 0.978          & 0.069          & 0.934        & \textbf{0.936} & \textbf{0.649} \\ \hline
model 4       & 0.974        & 0.976          & 0.069          & 0.933        & 0.935          & \textbf{0.649} \\ \hline
model 5       & 0.967        & 0.968          & 0.07           & 0.934        & \textbf{0.936} & \textbf{0.649} \\ \hline
\textbf{EnA}  & 0.98         & 0.98           & 0.069          & 0.934        & \textbf{0.936} & \textbf{0.649} \\ \hline
\textbf{EnM}  & 0.979        & 0.979          & 0.069          & 0.85         & 0.842          & 0.527          \\ \hline
\textbf{EnM2} & 0.979        & 0.978          & 0.069          & 0.933        & 0.935          & \textbf{0.649} \\ \hline
\textbf{HCNN} & 0.992        & \textbf{0.991} & \textbf{0.072} & 0.933        & 0.935          & \textbf{0.649} \\ \hline
\textbf{EnT}  & 0.99         & 0.97           & 0.104          & 0.332        & 0.27           & 0.093          \\ \hline
RF            & \textbf{1.0} & 0.971          & 0.068          & \textbf{1.0} & 0.718          & 0.366          \\ \hline
SVM           & 0.182        & 0.188          & 0.068          & 0.069        & 0.064          & 0.183          \\ \hline
LR            & 0.935        & 0.927          & 0.108          & 0.265        & 0.242          & 0.053          \\ \hline
\end{tabular}
\caption{Results for digit recognition experiments: (1) from MNIST as the source domain to SVHN as the target domain, (2) from SVHN as the source domain to MNIST as the target domain.}
\label{table2}
\end{table}
% ---------

% subsection: Hardware and software details
% \subsection{Hardware and software details}
% We used Python 3.8 as our programming language, and Keras\cite{chollet2015keras} as the framework for the CNNs and ensembles, and scikit-learn\cite{scikit-learn} for the traditional ML algorithms.
% ---------
\section{Results}
\Cref{table1,table2,table3,table4} show all the accuracy scores for every model on every problem. By analyzing the tables, we can notice the poor performance of the traditional ML models because they are being trained and tested on image datasets. However, the Random Forest model achieves high accuracy on the training set due to the fact that it is composed of many decision trees and can easily over-fit the training data, but we can see that when tested on the target domain we get very low accuracy. Moreover, while tuning the hyperparameters for the random forest, we noticed that the more we increase the number of trees the higher the training and validation accuracy until the training accuracy reaches 1.0, at which point the validation accuracy starts to plateau.
\begin{table}[h!]
\centering
\begin{tabular}{|c|c|c|c|c|c|c|}
\hline
Model         & \multicolumn{3}{c|}{USPS to MNIST}             & \multicolumn{3}{c|}{MNIST to USPS}             \\ \hline
              & $S_{train}$  & $S_{val}$      & $T$            & $S_{train}$  & $S_{val}$      & $T$            \\ \hline
model 1       & 0.996        & 0.976          & 0.776          & 0.998        & 0.994          & 0.968 \\ \hline
model 2       & 0.999        & 0.981          & 0.85           & 0.997        & 0.994          & 0.888          \\ \hline
model 3       & \textbf{1.0} & 0.98           & 0.794          & 0.996        & 0.993          & 0.958          \\ \hline
model 4       & \textbf{1.0} & 0.975          & 0.801          & 0.998        & 0.994          & 0.919          \\ \hline
model 5       & 0.999        & 0.981          & 0.859          & 0.996        & 0.993          & \textbf{0.973}          \\ \hline
\textbf{EnA}  & 0.999        & \textbf{0.982} & 0.852          & 0.998        & \textbf{0.995} & 0.962          \\ \hline
\textbf{EnM}  & \textbf{1.0} & \textbf{0.982} & 0.852          & 0.998        & \textbf{0.995} & 0.962          \\ \hline
\textbf{EnM2} & 0.999        & \textbf{0.982} & 0.864          & 0.998        & \textbf{0.995} & 0.957          \\ \hline
\textbf{HCNN} & \textbf{1.0} & 0.977          & \textbf{0.885} & 0.995        & 0.993          & 0.904          \\ \hline
\textbf{EnT}  & 0.999        & 0.942          & 0.112          & 0.962        & 0.945          & 0.113          \\ \hline
RF            & \textbf{1.0} & 0.941          & 0.098          & \textbf{1.0} & 0.968          & 0.118          \\ \hline
SVM           & 0.999        & 0.915          & 0.152          & 0.921        & 0.918          & 0.194          \\ \hline
LR            & 0.301        & 0.308          & 0.372          & 0.803        & 0.798          & 0.084          \\ \hline
\end{tabular}
\caption{Results for digit recognition experiments: (1) from USPS as the source domain to MNIST as the target domain, (2) from MNIST as the source domain to USPS as the target domain.}
\label{table3}
\end{table}

Another observation is that the CNN-based ensembles (EnA, EnM, EnM2) always give better accuracy in both domains across all the experiments, such as in CIFAR-to-STL (\Cref{table1}) where they reached 99\% accuracy in the training set and increased over the best individual model (of its base models) in the target domain by 2\% (from 66.4\% to 68.4\%). A similar outcome was observed in the SVHN-to-USPS experiment (\Cref{table4}).

On the other hand, we can notice a slight drop in accuracy in the ensemble compared to its best base model such as in the USPS-to-MNIST experiment (\Cref{table3}) where on the target domain the best performing model got 85.9\%, yet none of the ensembles got higher than that. This is because the other models in the ensemble have significantly less accuracy than the best model. However, the ensembles generally still have higher accuracy than the mean accuracy of their base models.

For some of the experiments, we do not see good generalization, such as in MNIST-to-SVHN experiment (\Cref{table2}), which is due to the huge domain gap between them. Even though the models achieve 95\%+ accuracy on training, they get very bad results on the target domain on testing, and in such cases, ensemble methods do not help much.

\begin{table}[h!]
\centering
\begin{tabular}{|c|c|c|c|c|c|c|}
\hline
Model         & \multicolumn{3}{c|}{USPS to SVHN}              & \multicolumn{3}{c|}{SVHN to USPS}             \\ \hline
              & $S_{train}$  & $S_{val}$      & $T$            & $S_{train}$  & $S_{val}$      & $T$            \\ \hline
model 1       & 0.998        & 0.974          & 0.115          & 0.957        & 0.952          & \textbf{0.707} \\ \hline
model 2       & 0.998        & 0.976          & 0.138          & 0.954        & 0.948          & 0.675          \\ \hline
model 3       & 0.999        & 0.974          & 0.11           & 0.957        & 0.947          & 0.714          \\ \hline
model 4       & 0.999        & 0.973          & 0.144          & 0.955        & 0.947          & 0.719          \\ \hline
model 5       & 0.998        & 0.972          & 0.123          & 0.951        & 0.952          & 0.737          \\ \hline
\textbf{EnA}  & 0.999        & 0.977          & 0.125          & 0.961        & \textbf{0.957} & \textbf{0.756} \\ \hline
\textbf{EnM}  & 0.998        & \textbf{0.981} & \textbf{0.159} & 0.961        & \textbf{0.957} & 0.755          \\ \hline
\textbf{EnM2} & 0.999        & 0.977          & 0.125          & 0.96         & \textbf{0.957} & 0.748          \\ \hline
\textbf{HCNN} & \textbf{1.0} & 0.979          & 0.08           & 0.985        & 0.962          & 0.604          \\ \hline
\textbf{EnT}  & 0.995        & 0.933          & 0.115          & 0.334        & 0.284          & 0.11           \\ \hline
RF            & \textbf{1.0} & 0.94           & 0.068          & \textbf{1.0} & 0.694          & 0.465          \\ \hline
SVM           & 0.994        & 0.947          & 0.148          & 0.124        & 0.125          & 0.167          \\ \hline
LR            & 0.301        & 0.308          & 0.068          & 0.261        & 0.239          & 0.06           \\ \hline
\end{tabular}
\caption{Results for digit recognition experiments: (1) from USPS as the source domain to SVHN as the target domain, (2) from SVHN as the source domain to USPS as the target domain.}
\label{table4}
\end{table}

% ---------

% subsection: ANN-En
% Table 1: All variants on all variants of problem 1
% Table 2: All variants on all variants of problem 2

% ---------

% subsection: Comparison with Deep Network
% Table 3: Accuracy and Training time of the best variant of ANN-En (from previous experiment) vs. the deep network 
% Scatter plots of the latent spaces

% ---------

% subsection: Comparison with ensemble model of traditional machine learning models
% Table 4: Similar to previous experiment but vs. Ensemble of [SVM, RF, LR] 

% ---------
\section{Conclusion}

By providing a different data augmentation for each base learner, we improved the generalization from a single source domain to an unseen target domain. Thus, this proved the usefulness of our ensemble approach, making it the simplest known method for domain generalization. Moreover, it can utilize weak models to get a more robust model. Additionally, note that the more base models there are, the more time it would need for training.

For future research, we can explore the effectiveness of the ensemble methods when using multiple source domains, how to use ensemble methods in domain adaptation, and how to best utilize the fact that we have access to the target domain. Also, we will explore how we can incorporate ensemble methods in current approaches for solving the domain adaptation and generalization problems.

\bibliographystyle{unsrt}
\bibliography{ref}
\end{document}